%% file: main.tex
\documentclass[letterpaper, 10 pt, conference]{ieeeconf}  

\IEEEoverridecommandlockouts                              

\input{preamble.tex}

\input{commands.tex}

\newcommand{\citet}[1]{\cite{#1}}
\newcommand{\citep}[1]{\cite{#1}}

\title{\LARGE \bf \titlelong}

\author{Ken Nakahara$^{1}$ and Roberto Calandra$^{1,2}$
\thanks{$^{1}$Learning, Adaptive Systems, and Robotics (LASR) Lab, TU Dresden, Dresden, Germany.
        {\tt\small \{knakahara, rcalandra\}@lasr.org}}%
\thanks{$^{2}$Center for Tactile Internet with Human-in-the-Loop (CeTI), Germany.}%
}

\begin{document}

\maketitle
\thispagestyle{empty}
\pagestyle{empty}


\begin{abstract}
	\input{0_abstract.tex}
\end{abstract}


\section{INTRODUCTION}
\label{sec:introduction}

	\input{1_introduction.tex}


\section{RELATED WORK}
\label{sec:related}

	\input{2_related.tex}


\section{HARDWARE SETUP}
\label{sec:hardware}

	\input{3_hardware.tex}


\section{ACTION-CONDITIONAL MULTI-MODAL MODELS FOR GENTLE GRASPING}  
\label{sec:approach}

	\input{4_approach.tex}


\section{DATA COLLECTION}
\label{sec:data_collection}

	\input{5_data_collection.tex}


\section{EXPERIMENTAL RESULTS}
\label{sec:result}

	\input{6_result.tex}


\section{CONCLUSION}
\label{sec:conclusion}

	\input{7_conclusion.tex}







\section*{ACKNOWLEDGMENT}

\input{99_acknowledgments.tex}



\bibliographystyle{IEEEtran}
\bibliography{main}

\end{document}

%% file: preamble.tex
\usepackage[english]{babel}

\usepackage{graphicx}
\usepackage{animate}
\usepackage{epsfig} 				
\usepackage{epstopdf}

\usepackage{listings}
\usepackage{color}
\usepackage{nameref}
\usepackage{hyperref}
\usepackage{amsmath}	 			
\usepackage{amssymb}  				

\usepackage{dsfont}			
\usepackage{mathtools}

\usepackage{epigraph}
\usepackage{lscape}
\usepackage[]{nomencl}				
\usepackage{algorithm}
\usepackage{algorithmic}
\usepackage{multicol}
\usepackage{multirow}
\usepackage{etoolbox}

\usepackage{caption}
\usepackage{subcaption}
\usepackage{wrapfig}

\usepackage{siunitx}

\usepackage{floatflt}

\usepackage{url}

\usepackage{bm}
\usepackage{xcolor}
\usepackage{colortbl}
\usepackage{hhline}

%% file: commands.tex
\newcommand{\email}[1]{\href{mailto:#1}{\nolinkurl{#1}}}

\renewcommand{\sec}[1]{Section~\ref{#1}}
\newcommand{\fig}[1]{Fig.~\ref{#1}}
\newcommand{\eq}[1]{Equation~\eqref{#1}}

\newcommand{\tab}[1]{Table~\ref{#1}}

\newcommand{\titlelong}[0]{Learning Gentle Grasping Using Vision, Sound, and Touch}

%% file: 0_abstract.tex
In our daily life, we often encounter objects that are fragile and can be damaged by excessive grasping force, such as fruits.
For these objects, it is paramount to grasp gently---not using the maximum amount of force possible, but rather the minimum amount of force necessary.
This paper proposes using visual, tactile, and auditory signals to learn to grasp and regrasp objects stably and gently.
Specifically, we use audio signals as an indicator of gentleness during the grasping, and then train an end-to-end action-conditional model from raw visuo-tactile inputs that predicts both the stability and the gentleness of future grasping candidates, thus allowing the selection and execution of the most promising action.
Experimental results on a multi-fingered hand over 1,500 grasping trials demonstrated that our model is useful for gentle grasping by validating the predictive performance (3.27\% higher accuracy than the vision-only variant) and providing interpretations of their behavior.
Finally, real-world experiments confirmed that the grasping performance with the trained multi-modal model outperformed other baselines (17\% higher rate for stable and gentle grasps than vision-only).
Our approach requires neither tactile sensor calibration nor analytical force modeling, drastically reducing the engineering effort to grasp fragile objects.
Dataset and videos are available at \href{https://lasr.org/research/gentle-grasping}{https://lasr.org/research/gentle-grasping}.

%% file: 1_introduction.tex
Grasping has been developed in modern robotics, but grasping fragile objects, such as fruits, with an appropriate amount of force remains challenging---a task we refer to as ``gentle grasping.''
Excessive grasping force can damage them, while insufficient force causes slippage or drop.

When such delicate robotic grasping is required, it is essential to consider the dynamic interaction with a target object.
To this end, introducing touch sensing is a natural demand, as tactile feedback has been proven crucial for humans in grasping \cite{johansson2009coding}.
However, due to the engineering costs of incorporating tactile sensors into conventional approaches, many grasping studies have been conducted with only vision sensing \cite{du2021vision}.
Although image-based algorithms have made progress \cite{li2021survey}, enabling fair grasping performance taking only visual feedback, relying solely on external cameras is prone to errors between the robot’s perception and the actual state.

Recent studies \cite{li2020review} show that touch sensing improves grasping performance, particularly with vision-based tactile sensors that capture local contact geometry as images. 
By processing tactile images, many works (e.g., \cite{calandra2017feeling, calandra2018more}) have effectively integrated visual and touch sensing and significantly improved grasping over vision alone.

Although multi-modal learning for grasping is promising, gentle grasping is still challenging.
Many studies have tackled this task due to industrial demand \cite{mandil2023tactile}, but they have often focused on hardware development, such as soft robotic hands \cite{shintake2018soft}.
Prior work \cite{calandra2018more} successfully learned to adjust a grasping force using the visuo-tactile model, but the target objects used are neither deformable nor fragile.

To train a policy for gentle grasping, a key question arises: how should the ``gentleness'' be defined?
A simple approach would optimize grasping force, as in \cite{calandra2018more} with a parallel jaw gripper, but it is not directly observable or controllable when using a multi-fingered hand, in general. 
In this work, we instead use sound as an indicator of gentleness, as many objects naturally provide audio cues when excessive force is applied.
However, the proposed framework can easily be extended to use, without loss of generality, any other measurable function of gentleness, such as the visual deformation of an object.
Since tactile feedback has been shown to improve grasping performance, incorporating auditory sensing is a promising avenue in the context of multi-modal learning, potentially offering perceptual insights beyond vision and touch.

\begin{figure}[t]
    \centering
    \includegraphics[width=1.0\linewidth]{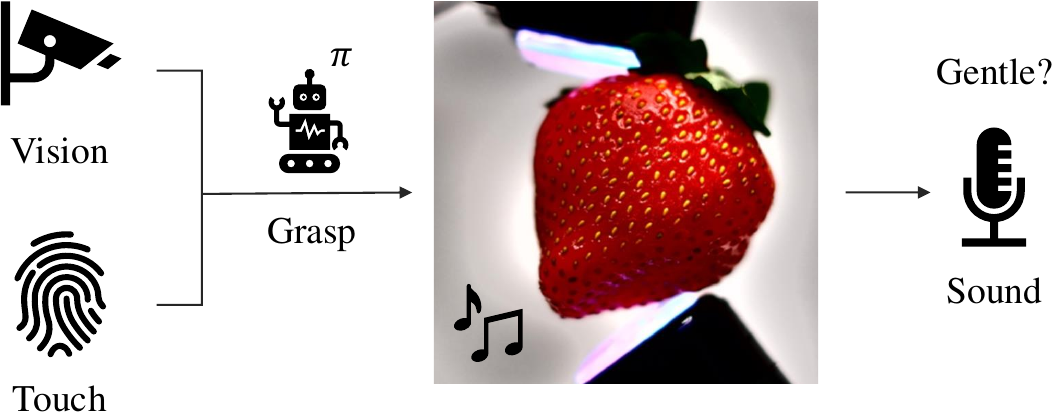}
    \caption{We propose a method for learning to grasp and regrasp fragile objects using visual, tactile, and audio signals. Based on the current visuo-tactile observations, the robot performs optimal grasping and evaluates grasp gentleness using sound produced during grasping. 
    We currently validate our method with a toy that emits sound under excessive grasping force.
    }
    \label{fig:vision_sound_touch}
\end{figure}

This work aims to realize gentle grasping by integrating auditory signals into the action-conditional model with vision and touch proposed in \cite{calandra2018more}, as illustrated in \fig{fig:vision_sound_touch}.
Unlike many studies with two-fingered grippers (e.g., \cite{calandra2018more, han2024learning}), we employ a multi-fingered hand, enabling more flexible and human-like grasping policies.
The proposed model utilizes visual input from an external camera and rich tactile feedback from DIGIT sensors \cite{lambeta2020digit} on the fingertips.
Additionally, auditory detection enables automatic labeling for the gentleness of a grasp in data collection.
Using a deformable toy that emits sound under excessive grasping force, real-world experiments demonstrate that our multi-modal model outperforms the vision-only variant and chance grasping regarding grasp stability and gentleness.
Trained end-to-end from raw visuo-tactile inputs, our approach needs neither calibration of tactile sensors nor analytical force modeling, drastically reducing the engineering effort for gentle grasping.

%% file: 2_related.tex
\subsection{Vision-based Tactile Sensors}
\label{sec:related_work_sensor}

Among various sensors for robotic dexterity, vision-based tactile sensors have made remarkable progress, using a deformable elastomer illuminated from within and an internal camera to capture contact interactions as high-resolution images \cite{shimonomura2019tactile}.
GelSight \cite{yuan2017gelsight} is one such sensor, capable of estimating normal and shear forces by analyzing the deformation of a grid of surface markers. 
DIGIT \cite{lambeta2020digit} refines GelSight with a more compact design, improved elastomer durability, and significantly lower production costs.

Since these optical sensors provide tactile reading as images, we can apply conventional image processing to them.
Over the past decade, convolutional neural networks (CNNs), which leverage spatial biases on images, have become the dominant approach in image processing \cite{li2021survey}.
The ease of using CNNs with vision-based tactile sensors facilitates multi-modal learning using vision and touch sensing.
As a result, various multi-modal models using these sensors have shown great grasping performance (e.g., \cite{calandra2017feeling, calandra2018more, han2024learning}).

\subsection{Learning for Gentle Grasping}
\label{sec:related_work_gentle_grasping}

Grasping is a core challenge in robotics, evolving from model-based methods \cite{shimoga1996robot} that typically analyze object geometry to determine optimal grasps, to model-free approaches \cite{kleeberger2020survey} using machine learning for generalized grasping policies.
More recently, end-to-end learning (e.g., \cite{calandra2018more,levine2018learning}) has emerged as the dominant paradigm for grasping.

Despite this progress, grasping fragile objects remains challenging, whereas high industrial demand \cite{mandil2023tactile}.
Although hardware developments, such as those in soft robotics \cite{shintake2018soft}, have enabled gentle grasping, they often require system-specific control theories and learning methods. 
Therefore, due to its cost-effectiveness and adaptability, integrating touch sensing into conventional rigid hands has been extensively explored.
\cite{romano2011human} is a pioneering study for gentle grasping using human-like tactile sensors with manual parameter tuning.
\cite{ford2023tactile} presents a force control scheme for a five-fingered robotic hand with omni-directional tactile sensors (TacTip \cite{ward2018tactip}).
By considering the similarity metrics of tactile images before and after grasping, the robot grasps 43 objects gently and applies the controller to human-to-robot handovers.
\cite{yamaguchi2017implementing} employs FingerVision, a vision-based tactile sensor with markers, to gently grasp objects, estimating force through marker displacements analyzed by predefined thresholds.

In contrast to these analysis-based methods, end-to-end learning approaches drastically reduce engineering effort.
However, they still struggle to learn the trade-off between grasp stability and gentleness.
\cite{han2024learning} proposes a Transformer-based grasping framework using visuo-tactile inputs to select appropriate forces by classifying fruit types and predicting grasping outcomes, achieving high success rates even for unseen fruits.
However, it optimizes force over only 13 discrete values and thus does not fully account for the physical interaction with objects.
\cite{nazari2023deep} uses the action-conditional tactile prediction and CNNs to control the contact position of a strawberry stem.
Similarly, \cite{huang2019learning} uses surprise-based intrinsic rewards in deep reinforcement learning to promote gentler multi-fingered touch.
Although these studies focus on pushing or touching rather than grasping, their methodology and motivation align with our work.
\cite{calandra2018more} optimizes grasping force with a gripper to a wide variety of target objects using the action-conditional model, but the target objects used are neither deformable nor fragile.
Our work extends this approach by incorporating sound to achieve more explicit learning for the grasp gentleness using a multi-fingered hand.

\subsection{Multi-modal Learning with Auditory Sensing in Robotics}
\label{sec:related_work_audio}

Although often overlooked in robotics compared to vision and touch, sound is crucial in our daily tasks with dynamic environmental interactions, as \cite{schaffert2019review} shows hearing sounds aids our motor learning.
In multi-modal learning, sound can complement vision and touch in robotic manipulation.
For instance, \cite{xu2022towards} integrates auditory, tactile, and proprioceptive inputs for robotic piano playing using reinforcement learning.
Another notable study \cite{liu2024maniwav} introduces an ear-in-hand device capturing synchronized audio-visual feedback for everyday tasks, employing a learning-from-demonstration (LfD) approach.
The proposed LfD framework highlights the potential for multi-modal learning with auditory sensing, and the processing method for audio data (e.g., adding artificial noise to the audio signal, encoding by an audio spectrogram transformer \cite{gong2021ast}) could be incorporated into our work.
Furthermore, \cite{jiang2024case} integrates three senses of vision, sound, and touch for cross-modal retrieval and shows the potential for improved robotic manipulation.

%% file: 3_hardware.tex
\begin{figure}[t]
    \centering
    \includegraphics[width=1.0\linewidth]{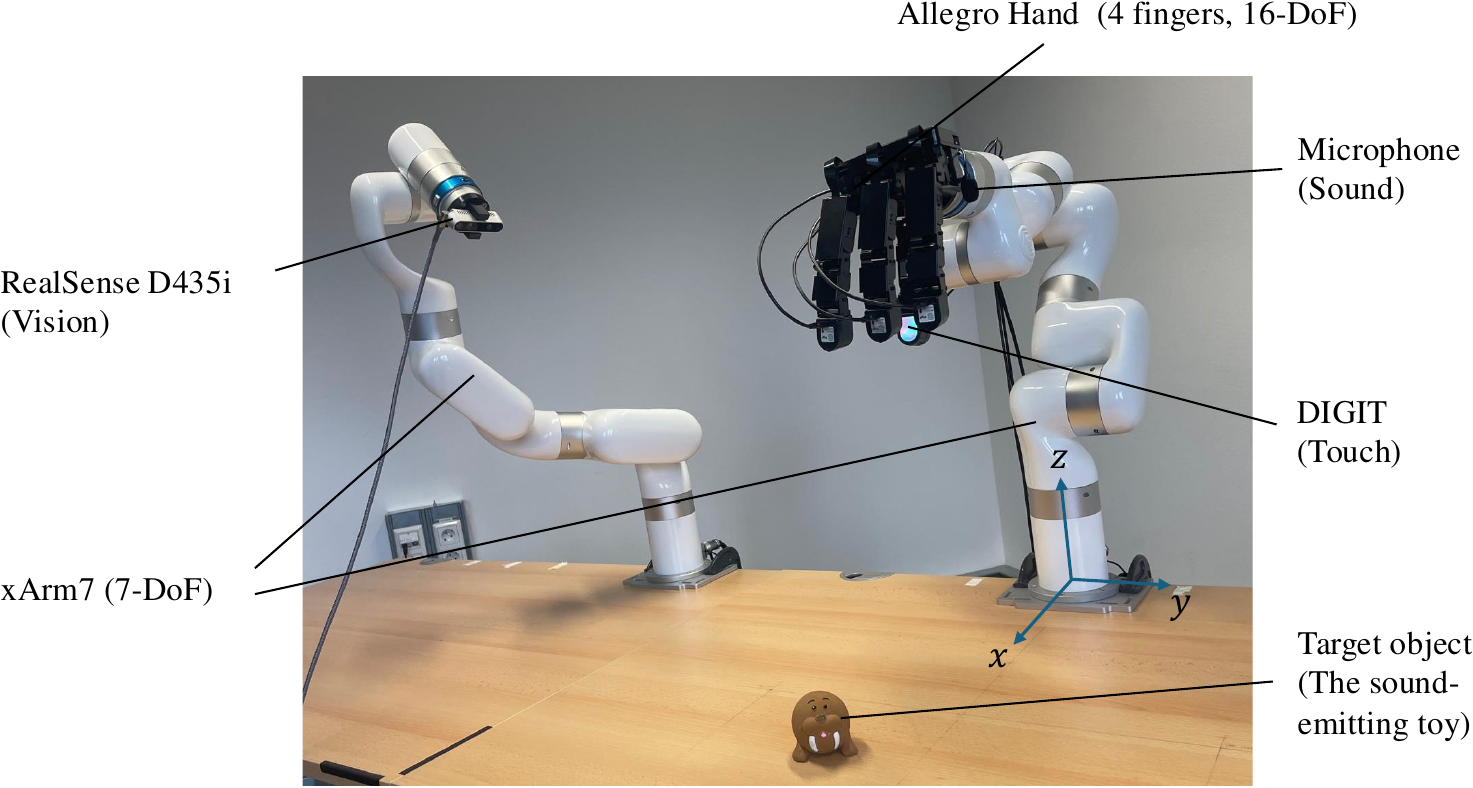}
    \caption{The experimental environment. The robot with the multi-fingered hand and two arms can record data from the RGB-D camera, DIGIT tactile sensors, microphone, and proprioception (e.g., joint angles) when grasping the object that makes a sound under excessive force. The world coordinates of this system are set at the base of the arm.
    }
    \label{fig:hardware}
\end{figure}

\begin{figure*}[t]
    \centering
    \includegraphics[width=1.0\linewidth]{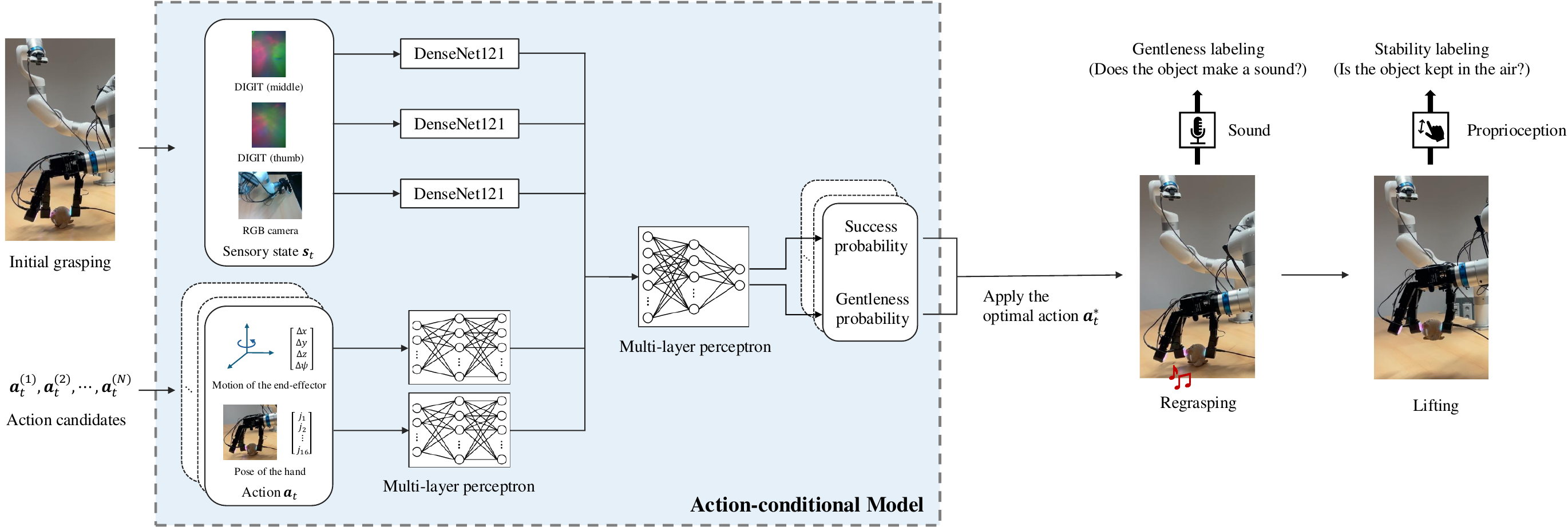}
    \caption{Overview of our learning control framework based on the action-conditional visuo-tactile model.
    Given the current visuo-tactile observations $\bm{s}_t$ and the action candidate $\bm{a}_t$, our model predicts the probabilities that the robot will grasp the object stably (without dropping it during lifting) and gently (without making a sound during regrasping $\bm{a}_t$).
    After processing the state-action pairs for all action candidates, the robot selects the best regrasp $\bm{a}^*_t$ among them based on these predictions.
    In the experiments, we also evaluate replacing DenseNet121 \cite{huang2017densely} with EfficientNetV2 B0 \cite{tan2021efficientnetv2} and ResNet50 \cite{he2016deep}.}
    \label{fig:architecture_model}
\end{figure*}

This work deals with sound, a challenging property to simulate.
Therefore, as shown in \fig{fig:hardware}, we conduct data collection and optimal control with trained models entirely in a real-world experimental environment.
Our robotic system consists of a four-fingered robotic hand (Allegro Hand ver. 4.0), two 7-DoF manipulators (xArm7), DIGIT sensors \cite{lambeta2020digit} at every fingertip, an external RGB-D camera (RealSense D435i), and a commodity microphone.
One of the arms has the hand as its end-effector, and the other is used for fixing the camera, capturing frames from a constant angle.
The Allegro Hand has four joints for each finger, i.e., 16-DoF in total.
For simplicity, we activate five joint angles on the thumb and the middle finger, and fix the other 11 joints to keep both fingers within the same plane.
This allows us to approximate the problem as the optimal control in two-dimensional kinematics.
The world coordinate system is set at the base of the arm with the hand, and the end-effector position is defined at the midpoint between the fingertips of the thumb and the middle finger.
As a visual sense of this system, RealSense provides RGB images at a resolution of $640\times480$ at \SI{30}{\hertz}.
DIGIT sensors, as a tactile sense, acquire RGB images of the contact surface at a resolution of $240\times320$ at \SI{60}{\hertz} over a sensing field of $16\times19$~\SI{}{\square\milli\meter}.
For audio detection, a standard commodity microphone is mounted on the side of the Allegro Hand, capable of recording audio data at a sampling rate of \SI{44.1}{\kilo\hertz}.
The target object is a deformable toy that produces a sound when subjected to a large amount of grasping force.
Given the relative sizes of the fingertips and the object, deviating from the object's center makes it harder to lift it stably, while applying excessive force to its center causes a distinct sound.
We selected this object from among the various candidates for its clear acoustic response when grasped.
However, as discussed in \sec{sec:force_prediction}, our experimental results indicate that environmental noise, such as sounds caused by actuators, can serve as important cues to evaluate grasp gentleness, suggesting the potential to generalize our approach to a variety of objects.
Our system integrates visual, tactile, auditory, and proprioceptive sensing, enabling multi-modal grasping.
All the robots and sensors used are cost-effective, many of them are open-sourced, and some components, such as the DIGIT and adapters, can even be self-built, making it easy to replicate and expand upon this work.

%% file: 4_approach.tex
Following \cite{calandra2018more}, we model grasping as a Markov Decision Process (MDP) to evaluate the stability of a future grasp using observations from the current grasp, enabling efficient adjustment for a more stable grasp.
We now present our learning control framework with an action-conditional model that also predicts grasp gentleness.

\subsection{End-to-End Outcome Prediction}
\label{sec:end2end_prediction}

Given the current sensory state $\bm{s}_t$ and the future action $\bm{a}_t$, we want the model to learn a function $\bm{f}(\bm{s}_t, \bm{a}_t)$ that predicts two probabilities: $f_s\in[0,1]$, the probability of grasping the object stably (without dropping during lifting), and $f_g\in[0,1]$, the probability of grasping it gently (without making a sound).
To this end, we parametrize the function $\bm{f}$ as a deep neural network.
As depicted in \fig{fig:architecture_model}, this model receives as input the sensory observation $\bm{s}_t$ during the current grasp, which consists of visuo-tactile images captured by the external camera and two DIGIT tactile sensors. 
Additionally, it takes the regrasp action $\bm{a}_t$ that specifies the relative displacement from the current end-effector position $\Delta x$, $\Delta y$, $\Delta z$, and orientation (yaw angle) $\Delta\psi$, as well as the absolute joint angles of the multi-fingered hand $(j_{1},\cdots,j_{16})^{\top}$.
Then, using the predicted probabilities $\bm{f}=(f_s, f_g)^{\top}$, the model makes binary predictions $\bm{\hat{o}}=(\hat{o}_s, \hat{o}_g)^{\top} \in \{0, 1\}^2$, where $\hat{o}_s$ represents a prediction for stability (with $\hat{o}_s = 1$ indicating a stable grasp), and $\hat{o}_g$ for gentleness (with $\hat{o}_g = 1$ indicating a gentle grasp).
To map from the probabilities $\bm{f}$ to the binary classifications $\bm{\hat{o}}$, we set the thresholds of 0.5 for stability and gentleness, respectively.

\subsubsection{Network Design}
\label{sec:network_design}
We employ DenseNet121 \cite{huang2017densely} to process visuo-tactile images $\bm{s}_t$, utilizing all layers up to its penultimate layer and replacing the final layer to the fully-connected layer for fine-tuning.
To identify the optimal CNN backbone for our work, we also explore the models with EfficientNetV2 B0 \cite{tan2021efficientnetv2} or ResNet50 \cite{he2016deep} instead of DenseNet121.
For these CNNs, all raw visuo-tactile images are first resized to $256\times256$, and then randomly cropped to $224\times224$ before processing in the network.
By this cropping, we augment the dataset twofold.
To highlight deformations, we also apply background subtraction by computing the difference between tactile images before and after contact.
The regrasp action $\bm{a}_t$, including the relative motion of the end-effector and the pose of the robot hand, is processed by multi-layer perceptrons (MLPs) consisting of two fully-connected layers, each with 1,024 hidden units, followed by Rectified Linear Unit (ReLU) activations and dropout with a rate of 0.25 for regularization in each layer.
To fuse these networks, the outputs of the five branches (RGB, two tactile images, and two action networks) are concatenated and passed through a two-layer fully-connected network. 
The first layer of this concatenated network contains 1,024 hidden units and is followed by a ReLU activation and dropout with a rate of 0.25.
The output layer, i.e., the head of this model, produces the grasp success and gentleness probabilities through sigmoid functions.

\subsubsection{Training}
\label{sec:training}
For model training, we collect a dataset of state-action-outcome tuples $(\bm{s}_i, \bm{a}_i, \bm{o}_i) \in X$ through a data collection process in which it performs random regrasp actions, described in \sec{sec:data_collection}.
Here, the grasp outcomes $\bm{o}_i$, i.e., true binary labels for grasp stability and gentleness, are automatically generated in a self-supervised manner.
The network $\bm{f}$ is then trained on this dataset $X$ by minimizing the binary cross-entropy (BCE) loss $L(\bm{f}, X)$.

For better predictive performance and faster training, we pre-train the CNNs using weights from a model trained to classify objects on ImageNet \cite{deng2009imagenet}.
For a learning rate, we employ the exponential scheduling \cite{li2019exponential}, promoting rapid convergence in the initial training stages while allowing for stable learning in later epochs to prevent overfitting.

\subsection{Regrasp Optimization Using Trained Models}
\label{sec:regrasp_optimization}

Our goal is to execute the stable and gentle grasp using the trained model $\bm{f}$.
In \cite{calandra2018more}, the action-conditional model is trained to predict the success probability $f_s$ of a future grasp, and select the optimal action $\bm{a}_t^*$ with the highest expected probability in a set of $N$ action candidates; $A = \{\bm{a}_t^{(j)}\}_{j=1}^{N}$.
In contrast, our model can also output the grasp gentleness probability $f_g$, and thus we can solve the constrained optimization problem for gentle grasping as follows:
\begin{equation}
\label{eq:constrained_optimization}
    \bm{a}_t^*=\arg \max_{\bm{a} \in A} f_s(\bm{s}_t, \bm{a}) \quad \text{s.t.} \quad f_g(\bm{s}_t, \bm{a}) > T_{\text{gentle},}
\end{equation}
where $T_{\text{gentle}}$ is a threshold for the gentleness probability.
To solve \eq{eq:constrained_optimization}, we perform a random search in the action space. 
We first randomly sample action candidates and predict their success and gentleness probabilities using the learned model $\bm{f}$.
Then, we select the action with the highest success probability $f_s$ among those whose gentleness probability $f_g$ exceeds the threshold $T_{\text{gentle}}$.

%% file: 5_data_collection.tex
\begin{figure}[t]
    \centering
    \includegraphics[width=1.0\linewidth]{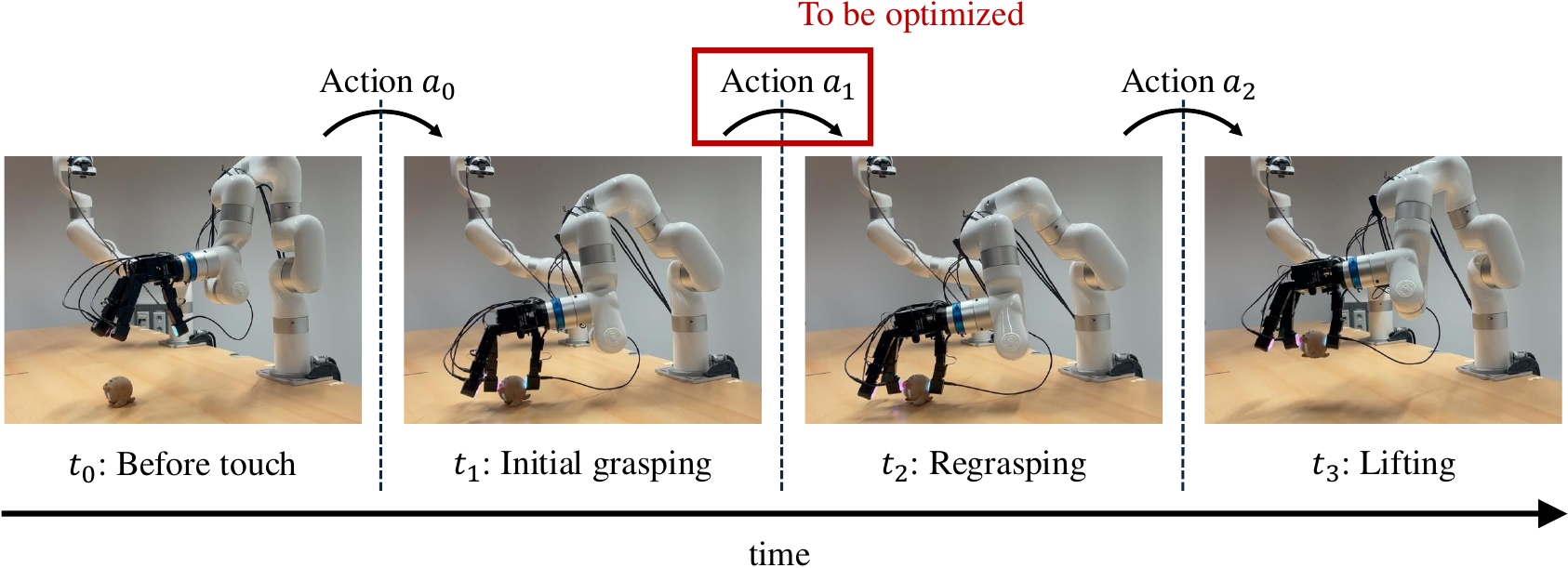}
    \caption{Overview of the data collection process. Through the four phases shown in the figure, visuo-tactile images and audio data are stored, a regrasp action $\bm{a}_1$ is randomly sampled, and grasp stability and gentleness labels for the obtained sample are automatically generated.}
    \label{fig:data_collection}
\end{figure}

\textbf{Data Collection Process:}
To collect the training data, we designed a self-supervised automatic data collection process.
This process consists of four phases: before grasping, initial grasping, regrasping, and lifting, as shown in \fig{fig:data_collection}.
Through the process, it automatically gives grasp outcomes, i.e., two binary labels indicating grasp stability and gentleness.

In the first phase, the object is placed at an arbitrary position within $x \in [450, 650]$ and $y \in [-200, 200]$ in the world coordinates ($200\times400$~\SI{}{\square\milli\meter}).
Then, the position of the object $(x, y)$ is detected by processing depth data from RealSense.
This detection can be achieved by selecting the point with the highest difference in depth value between before and after setting the object within the region of interest.
Afterward, the end-effector is positioned at the height $z=300$~\SI{}{\milli\meter}, above the object's height, with the hand opened wider than the object's width.
The index and ring fingers, unused in this experiment, remain fixed in subsequent processes.

Next, for initial grasping, the end-effector's orientation (yaw angle $\psi$) is randomly set within $\psi \in [-30^\circ,30^\circ]$ and then its position is set to the detected object's position $(x, y)$ with small perturbations introduced for data diversity.
After moving above the object, the height is sampled from $ z \in [5, 30]$, where this range is selected to ensure contact with the object but without conflicting with the ground.
The two fingers then close to pre-set poses, applying enough force to alter tactile readings, and sensory data (RGB, depth, and two tactile images) are recorded during this grasping.

We then sample a regrasp action $\bm{a}_1$, which includes the relative end-effector's motion $(\Delta x, \Delta y, \Delta z, \Delta \psi)^\top$ and the absolute joint angles of the hand $(j_1,j_2,\cdots,j_{16})^\top$.
The translations $(\Delta x, \Delta y, \Delta z)$ are randomly selected from $[-10,10]$, respectively, and rotation $ \Delta \psi$ from $[-15^\circ,15^\circ]$.
The hand pose is sampled between the pre-defined minimum pose that always fails to grasp and the maximum pose that consistently produces a loud sound, with the five active joints sampled within a range of \SI{0.3}{\radian} ($\approx 17.2^\circ$).
The sampled regrasp $\bm{a}_1$ is then executed while recording sound for \SI{3}{\second}, assigning a gentleness label: 0 (non-gentle) if the sound exceeds a pre-set threshold; otherwise, 1 (gentle).
This auditory threshold was manually determined through multiple grasp trials to ensure that a non-gentle label is assigned when the object produces a distinct sound.
We confirmed that, in this setting, even when the object itself does not make such a sound, the sounds caused by actuators or collision under excessive grasping are detected as non-gentle, while unrelated noise (e.g., human speech near the experimental setup) does not affect the detection.

After regrasping, the robot attempts to lift the object and hold it in the air.
If the object remains within the fingers for \SI{4}{\second} without dropping, it is labeled with 1 (success); otherwise, 0 (failure).
To automate this labeling, we train a logistic regression model to learn the relationship between grasp success labels and actual fingertip displacements during grasping, calculated via kinematics from observed joint angles of the fingers.
This model is first trained on supervised data and then progressively fine-tuned during the data collection using all the available data.

Lastly, the robot resets for the next trial.
After a failed grasp, it simply returns to the initial position.
If successful, it places the object at random within $x \in [450, 650]$ and $y \in [-200, 200]$ before returning to the start.

\textbf{Collected Data:}
We collected 1,500 grasping samples over \SI{25}{\hour} (about \SI{1}{\minute} per trial) with labels shown in \tab{data_labels}.
While most labels were generated through self-supervision, some were manually corrected based on visual inspection.
As shown in \tab{data_labels}, stability labels are balanced, but gentleness labels remain skewed due to the rarity of failure and non-gentle cases, despite manual tuning of the sound threshold.

To augment our dataset, we utilized sensory observations at two moments: after releasing the initial grasp and during regrasping.
The first moment forms a state-action pair with the regrasp action $\bm{a}_1$, while the second creates a pair with the current hand pose and zero end-effector movement.
Plus, the cropping method (see \sec{sec:network_design}) doubled the dataset.
In total, we expanded its size sixfold, reaching 9,000 samples.

\begin{table}[t]
    \centering
    \caption{Generated labels of collected data from 1,500 grasping trials.
    Each sample is annotated with two binary labels: grasp stability (success or failure) and grasp gentleness (gentle or non-gentle).
    Gentleness labels are not balanced due to the inherent rarity of failure and non-gentle $(0,0)$ cases.}
    \label{data_labels}
    \renewcommand{\arraystretch}{1.15}
    \begin{tabular}{l|l|c|c|c}
        \multicolumn{2}{c}{}&\multicolumn{2}{c}{Gentleness}&\multicolumn{1}{c}{}\\
        \cline{3-4}
        \multicolumn{2}{c|}{}&1 (gentle) & 0 (non-gentle) & \multicolumn{1}{c}{Total}\\
        \cline{2-4}
        \multirow{2}{*}{Stability}& 1 (success) & 364 & 385 & 749\\
        \cline{2-4}
        & 0 (failure) & 609 & 142 & 751\\
        \cline{2-4}
        \multicolumn{1}{c}{} & \multicolumn{1}{c}{Total} & \multicolumn{1}{c}{973} & \multicolumn{1}{c}{527} & \multicolumn{1}{c}{1500}\\
    \end{tabular}
\end{table}

%% file: 6_result.tex
We now validate our model and its variations by comparing their predictive performance. 
Following, we analyze the acquired behavior of the learned model.
Lastly, real-world experiments demonstrate that the trained multi-modal model outperforms other baselines for stable and gentle grasping.

\subsection{Model Evaluation}
\label{sec:model_evaluation}

\begin{figure}[t]
    \centering
    \includegraphics[width=1.0\linewidth]{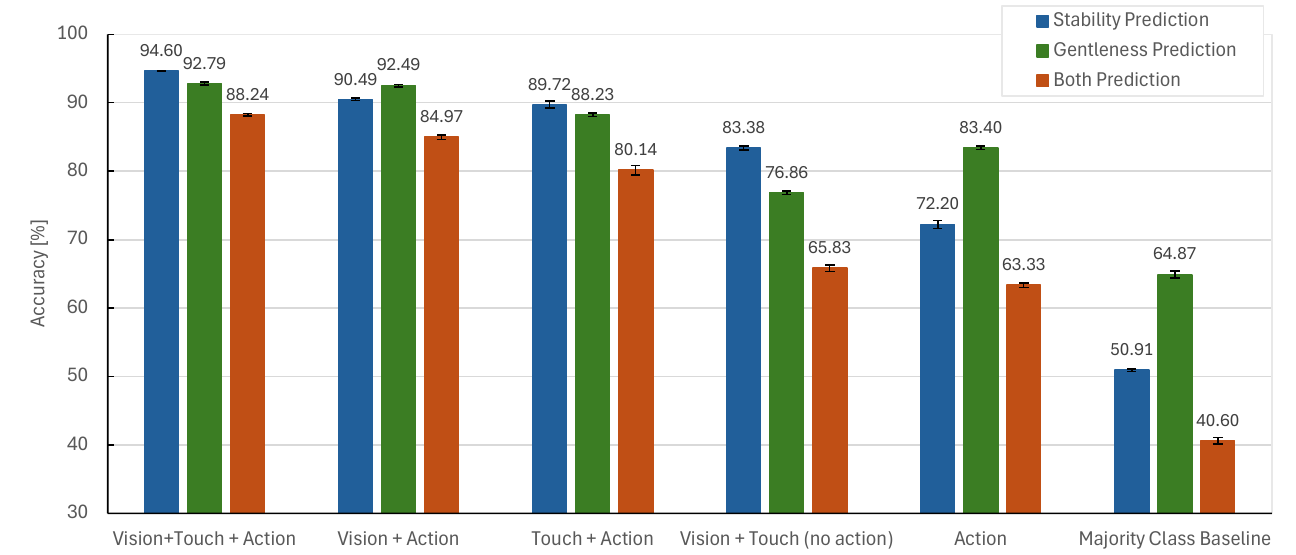}
    \caption{Mean accuracies of the models with different modalities for grasp stability and/or gentleness predictions in 5-fold cross-validation using 9,000 samples, with standard errors represented by error bars. 
    The accuracy for both predictions (orange bars) shows the proportion of the cases where both stability and gentleness are predicted correctly.
    The multi-modal model with vision, touch, and action inputs showed the highest performance for both predictions.
    }
    \label{fig:multi_modal_accuracy}
\end{figure}

\begin{figure}[t]
    \centering
    \includegraphics[width=1.0\linewidth]{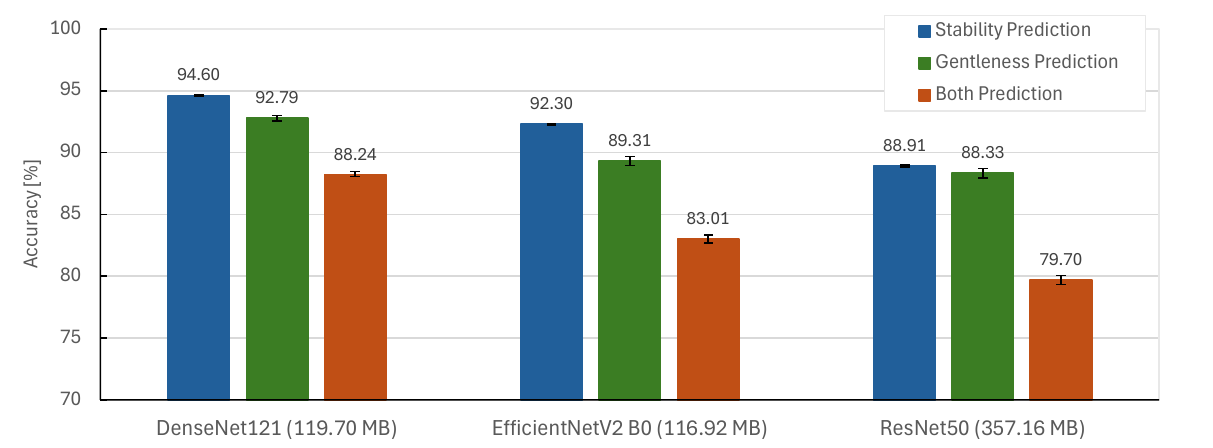}
    \caption{Mean accuracies of the multi-modal models with different CNN backbones for grasp stability and/or gentleness predictions in 5-fold cross-validation using 9,000 samples, with standard errors and model sizes (the memory usage when model parameters are stored as 4 bytes).
    The model with DenseNet121 showed the highest predictive accuracies, followed by EfficientNetV2 B0 and ResNet50.
    }
    \label{fig:diff_cnn_accuracy}
\end{figure}

We first examined whether the proposed model can learn to correctly predict future grasping outcomes as two binary classifications.
To assess the contribution of each modality, we also prepared models with reduced inputs, such as those excluding tactile networks.
We evaluated each model by 5-fold cross-validation with 9,000 samples.
For all models, training was conducted with a batch size of 8 for 500 epochs and the initial learning rate of $10^{-4}$ with a decay factor of 0.98 for exponential scheduling.
Note that the values of these hyperparameters were selected through our extensive tuning.

\fig{fig:multi_modal_accuracy} shows the accuracies of models with different modalities, all adopting DenseNet121 as the backbone, in the 4-class classification.
These results confirmed that the model with all modalities (``Vision + Touch + Action’’) achieved the highest accuracy (88.24\%) among various modality configurations.
Although other models with action inputs underperformed compared to this best model, they significantly outperformed the model without actions (``Vision + Touch (no action)’’). 
This model is equivalent to the unconditional model used in \cite{calandra2017feeling}, confirming that our action-conditional model effectively assessed the impact of different actions.
Furthermore, comparing the model without sensory inputs (``Action’’) to the model without actions (``Vision + Touch (no action)’’), we found that the action inputs significantly contributed to predicting grasp gentleness, as the pose of the robotic hand is directly related to the grasping force. 
Meanwhile, removing tactile networks from the best model (see ``Vision + Action’’) mainly impacted stability prediction, suggesting that touch provides critical information about grasp stability that vision alone could not capture.
Overall, all models significantly outperformed the baseline that simply predicted the majority class in each validation set.
We also validated the models with all modalities employing different CNN backbones: DenseNet121 \cite{huang2017densely}, EfficientNetV2 B0 \cite{tan2021efficientnetv2}, and ResNet50 \cite{he2016deep}.
As shown in \fig{fig:diff_cnn_accuracy}, the model with DenseNet121 had the highest predictive accuracies, followed by EfficientNetV2 B0 and ResNet50.
We assume that the feature reuse mechanism of DenseNet, where each layer receives outputs from all preceding layers, contributed to capturing intricate details on tactile images.

\subsection{Understanding the Learned Model}
\label{sec:understanding_model}

Since our model was trained end-to-end without manual design for its behavior, we analyzed its learned policies.
Here, we focused on the best-performing multi-modal model with DenseNet121 and examined the result from the iteration with the highest accuracy in the cross-validation with 9,000 samples (7,200 for training and 1,800 for validation).

\subsubsection{Confusion Matrix}
\label{sec:confusion_matrix}

We first inspected the confusion matrix in \tab{confusion_matrix} for grasp outcome predictions with four classes (stability: 1/0, gentleness: 1/0).
The diagonal elements represent the correct predictions, giving an overall accuracy of 88.83\%.
Although the performance of the model was mostly balanced between classes, it showed a slightly higher misclassification rate for the $(0,0)$ case, where the object produced a sound and was not lifted.
This stemmed from the inherent rarity of such a case (see \tab{data_labels}), making it harder for the model to learn to correctly classify that case.

\begin{table}[t]
    \centering
    \caption{Confusion matrix for (stability: 1/0, gentleness: 1/0) predictions by our best-performing model.
    This table represents the two binary classifications on 1,800 validation samples at the iteration with the highest accuracy (88.83\%) during 5-fold cross-validation.
    Due to the inherent rarity of failure and non-gentle cases, the model had a slightly lower accuracy on the $(0,0)$ class than others.}
    \renewcommand{\arraystretch}{1.15}  
    \begin{tabular}{>{\centering\arraybackslash}p{0.7cm}|>{\centering\arraybackslash}p{0.5cm}|>{\centering\arraybackslash}p{0.5cm}|>{\centering\arraybackslash}p{0.5cm}|>{\centering\arraybackslash}p{0.5cm}|>{\centering\arraybackslash}p{0.5cm}|>{\centering\arraybackslash}p{0.6cm}>{\centering\arraybackslash}p{1.0cm}}
        \multicolumn{2}{c}{}&\multicolumn{4}{c}{Predicted}&\multicolumn{1}{c}{}\\
        \hhline{~~|-|-|-|-|~}
        \multicolumn{2}{c|}{}&(1,1)&(1,0)&(0,1)&(0,0)&\multicolumn{1}{c}{Total}&\multicolumn{1}{c}{Acc. [\%]}\\
        \hhline{~|-|-|-|-|-|~}
        & (1,1) & \cellcolor{red!15}388 & 25 & 25 & 3 & 441 & 87.98 \\
        \hhline{~|-|-|-|-|-|~}
        \multirow{2}{*}{Actual}&(1,0)& 42 & \cellcolor{red!15}405 & 4 & 4 & 455 & 89.01\\
         \hhline{~|-|-|-|-|-|~}
        & (0,1) & 37 & 7 & \cellcolor{red!15}675 & 13 & 732 & 92.21\\
         \hhline{~|-|-|-|-|-|~}
        & (0,0) & 5 & 10 & 26 & \cellcolor{red!15}131 & 172 & 76.16\\
         \hhline{~|-|-|-|-|-|~}
        \multicolumn{1}{c}{} & \multicolumn{1}{c}{Total} & \multicolumn{1}{c}{472} & \multicolumn{1}{c}{447} & \multicolumn{1}{c}{730} & \multicolumn{1}{c}{151} & \multicolumn{1}{c}{1800} & \multicolumn{1}{c}{88.83}\\
    \end{tabular}
    \label{confusion_matrix}
\end{table}

\subsubsection{Relationship between Grasping Force and Model Prediction}
\label{sec:force_prediction}

\begin{figure}[t]
    \centering
    \includegraphics[width=1.0\linewidth]{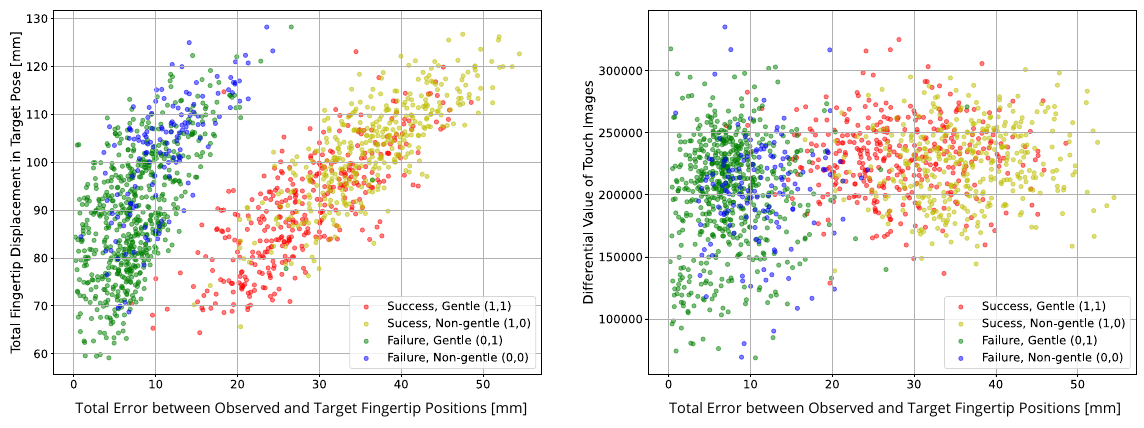}
    \caption{Scatterplots between fingertip displacements in target finger poses and position errors of fingertips (left), and between differential pixel values of tactile images and the position errors (right) for 1,500 grasping samples assigned four-class grasp labels.
    Each force metric represents the summed value for both fingers.
    We concluded that the fingertip position error is the best force alternative considering its effectiveness in distinguishing grasp stability and gentleness. 
    }
    \label{fig:force_alternatives}
\end{figure}

We further explored how the learned model predicted grasp success and gentleness probabilities concerning applied force. 
Since direct force measurement is impractical with our multi-fingered hand, we explored three alternative grasping force representations: (1) fingertip displacement---the shift of target fingertip positions before and after grasping, based solely on target joint angles; (2) fingertip position error---the deviation between observed and target fingertip positions in post-grasp, considering object interaction and external forces; (3)  tactile image differentials between pre- and post-grasp.
We examined their relationship with grasp stability and gentleness labels in 1,500 grasping trials.

First, the left of \fig{fig:force_alternatives} plots the relation between (1) fingertip displacements and (2) position errors, forming two distinct clusters in terms of grasp success.
In the right cluster with nearly all success labels, the two metrics strongly correlate, and higher values of both metrics correspond to an increased proportion of non-gentle grasps.
This indicates that these two metrics effectively represent the grasping force when both fingers make firm contact with the object.
In the left cluster with mostly failure labels, the two metrics show little correlation in the region with low metric values, where insufficient force was applied to lift the object.
However, when both metrics take moderately large values---where the fingers may have collided---a correlation emerges, and non-gentle grasps increase with higher values.
Notably, this implies that environmental noise, largely from the robot itself or the collision, aided in classifying grasp gentleness even when the object was not grasped.
Analyzing each metric separately, the position error more clearly distinguishes stability labels, while both metrics can differentiate gentleness labels at extremely low or high values.
Overall, these proprioception-based metrics have been validated as effective grasping force alternatives, with the position error metric proving particularly reliable for classifying grasp stability.

The right of \fig{fig:force_alternatives} illustrates the relation between (3) the differential tactile pixel values and (2) the position errors, showing little correlation between these metrics.
Below the value of 150,000, grasp failure is distinguishable, indicating barely contact with the object, but classification remains difficult elsewhere.
Thus, differential tactile values proved unsuitable for grasp force representation in this experiment.

\begin{figure}[t]
    \centering
    \includegraphics[width=1.0\linewidth]{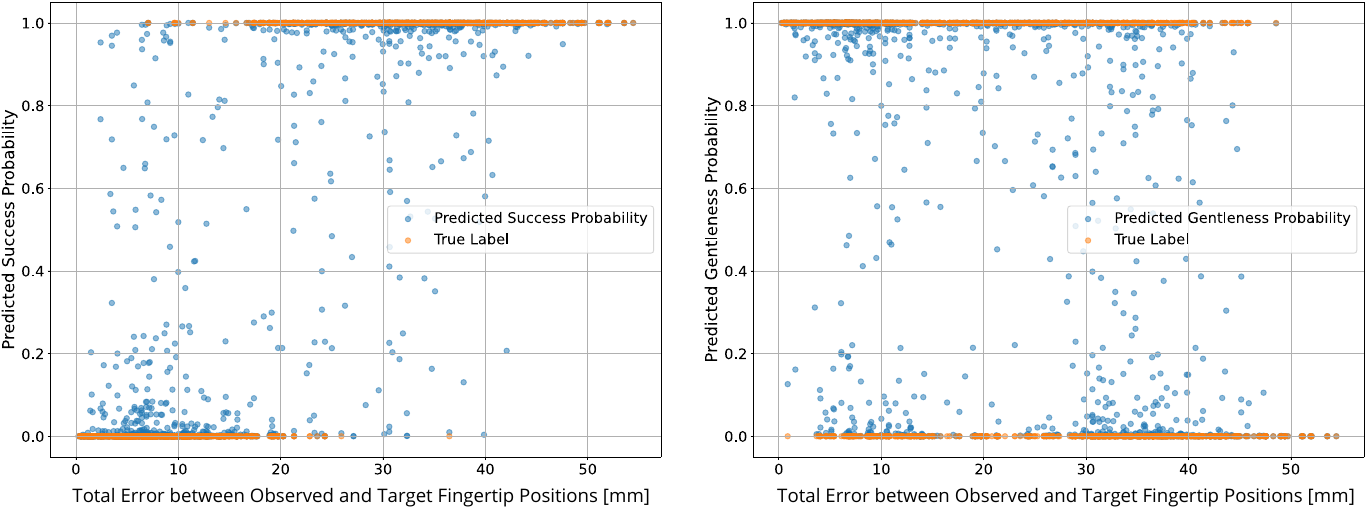}
    \caption{Scatterplots between the force metric (total position error of two fingertips) and the predicted success probability (left) and gentleness probability (right) for 1,800 validation samples.
    The model learned the tradeoff between grasp stability and gentleness, observing that greater applied force enhances stability while diminishing gentleness.}
    \label{fig:relation_error_preds}
\end{figure}

Through the examination above, we adopted the most suitable force alternative (i.e., fingertip position error) for analyzing the model's behavior.
\fig{fig:relation_error_preds} visualizes the relation between this force alternative and the model's predictions for grasp success and gentleness probabilities on 1,800 validation samples. 
The results showed that as the force metric increased, the trained model predicted a higher grasp success probability, while the predicted gentleness probability decreased.
These behaviors align with our intuitive expectations, suggesting that the trained model can effectively guide the selection of an appropriate grasping force.

\subsection{Evaluation of Gentle Grasping}
\label{sec:gentle_grasping}

\begin{figure}[t]
    \centering
    \includegraphics[width=1.0\linewidth]{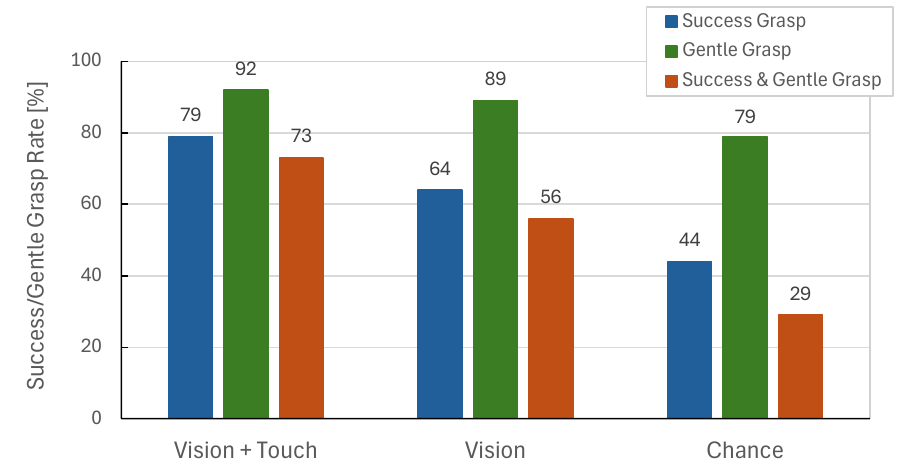}
    \caption{The rates for success and/or gentle grasps using the trained models with different modalities in 100 real-world grasping tests.
    The multi-modal model outperformed the model without tactile modality and chance grasping regarding both grasp stability and gentleness.
    }
    \label{fig:optimal_control}
\end{figure}

Finally, we tested the DenseNet121-based model trained with 9,000 samples on the real robot.
In this experiment, the robot attempted to gently grasp and stably lift the object using the models. 
Each test began by identifying the object's position with perturbations and randomly sampling the height and orientation of the end-effector for the initial grasp with a fixed hand pose, the same as the data collection procedure.
Since the robot used a camera to identify the object's position and initiated the grasping process without touch, a fair comparison for models without visual modality was not feasible.
We thus evaluated grasping performance using models with and without tactile networks, as well as a random regrasp approach.
For action optimization, we considered the same sampling range for the relative motion of the end effector and the five active finger joints as in the data collection (see \sec{sec:data_collection}).
From this action space, 95,000 random actions were sampled, and an additional 5,000 actions were sampled where only finger joint angles were varied and the end-effector's motion was set to zero.
Using the trained models, we solved the constrained optimization problem in \eq{eq:constrained_optimization} for all action candidates with a gentleness threshold $T_\text{gentle}=0.95$ (95\%).
Under this condition, if the predicted grasp success probability of the best candidate also exceeded 95\%, the future action included lifting the object.
Otherwise, we solved an unconstrained optimization problem for only grasp stability to determine the next end-effector motion for the following grasp.
This grasp used the same hand pose as the initial grasp, and a new set of actions was sampled to solve the constrained optimization problem again.

\fig{fig:optimal_control} shows the results of 100 grasping tests for each model. 
Since the object's orientation would affect its position detection and, in turn, grasping success, we conducted the first 50 trials with it upright and the remaining 50 with it lying down.
Overall, our models significantly outperformed the random grasp approach.
We also confirmed that while the vision-only variant underperformed the multi-modal model, it still exhibited high performance in gentle grasps, which aligns with the validation results in \sec{sec:model_evaluation}.
Comparing the gentle grasp rate of the chance grasps with the labels generated during data collection (see \tab{data_labels}), we observed its increase. 
We attributed this to a repair for the looseness of the finger bases that occurred after collecting all the data.

%% file: 7_conclusion.tex
This paper proposed an end-to-end learning control framework for gentle grasping. 
Our action-conditional model explicitly predicts both the stability and gentleness of a future grasp from visuo-tactile images, employing a multi-fingered hand with DIGIT tactile sensors.
We trained the models on data from 1,500 grasping trials, where automatic labeling was conducted for grasp stability via the robot's proprioception and gentleness via auditory sensing. 
Validations for a variety of our models demonstrated the effectiveness of different modalities and network architectures, and confirmed that our multi-modal model, integrating vision, touch, and action inputs, significantly outperformed other baseline models in predicting grasp outcomes.
We also explored grasping force representations and concluded that fingertip position error is the best force metric with a multi-fingered hand.
The analysis of the relationship between this force metric and the model's predictions demonstrated that the model correctly learned the trade-off between grasp stability and gentleness.
Finally, real-world experiments confirmed the superior performance of our multi-modal model over baselines, achieving higher success and gentle grasp rates (17\% higher than the vision-only model and 44\% higher than chance grasping).

Our current approach has several limitations that could be relaxed in future work:
First, we used proprioception for labeling grasp stability, which is ad hoc for our object's properties.
However, we could make it object-independent using tactile images as in \cite{calandra2018more} by training a CNN on larger tactile data to cope with subtle tactile differences in gentle grasp and using proprioception to filter collision cases.
Second, we employed a manually determined auditory threshold as an indicator of grasp gentleness. 
Incorporating more expressive functions, potentially learned from audio spectrograms or even multi-modal data, could enable a more adaptive and robust measurement of gentleness.
Lastly, integrating Digit360~\cite{lambeta2024digitizing}, an advanced multi-modal tactile sensor, offers richer audio and tactile readings on the contact surface, enhancing the quality of grasping and even lifting.

%% file: 99_acknowledgments.tex
This work is partly supported by the project ``Genius Robot'' (01IS24083) funded by the Federal Ministry of Education and Research (BMBF), by the German Research Foundation (DFG, Deutsche Forschungsgemeinschaft) as part of Germany’s Excellence Strategy – EXC 2050/1 – Project ID 390696704 – Cluster of Excellence “Centre for Tactile Internet with Human-in-the-Loop” (CeTI) of Technische Universität Dresden by Bundesministerium für Bildung und Forschung (BMBF), and by the German Academic Exchange Service (DAAD) in project 57616814 (\href{https://secai.org/}{SECAI} \href{https://secai.org/}{School of Embedded and Composite AI}).
We also thank the ZIH at TU Dresden for providing computing resources.